\pdfoutput=1

\documentclass[11pt]{article}

\usepackage{acl}
\usepackage{graphicx}
\usepackage{booktabs}
\usepackage{subfig}
\usepackage{times}
\usepackage{latexsym}

\usepackage[T1]{fontenc}

\usepackage[utf8]{inputenc}

\usepackage{microtype}

\title{Isomorphic Cross-lingual Embeddings for Low-Resource Languages}

\author{Sonal Sannigrahi\textsuperscript{1,2} Jesse Read\textsuperscript{2} \\
 \textsuperscript{1}Saarland University \\
  \textsuperscript{2}École Polytechnique \\
    \texttt{sosa00001@stud.uni-saarland.de}  \texttt{jesse.read@polytechnique.edu} \\}

\begin{document}
\maketitle
\begin{abstract}
Cross-Lingual Word Embeddings (CLWEs) are a key component to transfer linguistic information learnt from higher-resource settings into lower-resource ones. Recent research in cross-lingual representation learning has focused on offline mapping approaches due to their simplicity, computational efficacy, and ability to work with minimal parallel resources. However, they crucially depend on the assumption of embedding spaces being approximately isomorphic i.e. sharing similar geometric structure, which does not hold in practice, leading to poorer performance on low-resource and distant language pairs. In this paper, we introduce a framework to learn CLWEs, without assuming isometry, for low-resource pairs via joint exploitation of a related higher-resource language. In our work, we first pre-align the low-resource and related language embedding spaces using offline methods to mitigate the assumption of isometry. Following this, we use joint training methods to develops CLWEs for the related language and the target embedding space. Finally, we remap the pre-aligned low-resource space and the target space to generate the final CLWEs. We show consistent gains over current methods in both quality and degree of isomorphism, as measured by bilingual lexicon induction (BLI) and eigenvalue similarity respectively, across several language pairs: \{Nepali, Finnish, Romanian, Gujarati, Hungarian\}-English. Lastly, our analysis also points to the relatedness as well as the amount of related language data available as being key factors in determining the quality of embeddings achieved. 

\end{abstract}

\section{Introduction}

In a world with over 7000 spoken languages, out of which nearly 43\% are endangered, there is an acute need for accurate language technology systems that ensure equal access of resources in a predominantly digital world \footnote{\url{http://www.unesco.org/languages-atlas/en/statistics.html}}. Early successes in neural natural language tasks were primarily data-driven and English focused \cite{belinkov2019analysis}, however as we move on to low-resource, multi-lingual scenarios it becomes imperative to develop meaningful representations. 

\noindent Taking machine translation (MT) as an example,
we have observed remarkable progress over the last few years propelled by advances in neural language modelling. This success has been mainly confined to major world languages \cite{hassan2018achieving, liu2020very}, however, a significant proportion of languages are endangered or otherwise have a very scarce amount of digital resources which presents serious challenges for training MT systems. Rather than traditional expert-guided feature engineering, neural MT (NMT), like deep neural architectures more generally, require notoriously large data sets from which to extract features automatically in the context of hidden layers; for example with recurrent \cite{cho2014properties, schmidhuber1997long}, and attention mechanisms \cite{bahdanau2014neural}. It is for this reason that the most impressive results (e.g., \cite{liu2020very,barrault-etal-2019-findings}) come from languages with large scale digital resources such as parallel corpora with which to train them. This is, however, not the case for most minority languages. 

\noindent Current research in expanding natural language tools for low-resource settings has focused on transferring information from languages for which we have sufficient data to model correctly \cite{chen-etal-2019-multi-source, lample2018phrasebased, lample2017unsupervised}. In order to achieve this, cross-lingual word embeddings (CLWEs) are important which is what we focus on in this work. As CLWEs represent words from multiple languages in a shared vector space, they are key in promoting language sharing across low and high-resource languages. The two primary approaches in learning CLWEs are: 1) \textit{mapping methods} which independently map monolingual word embeddings by learning a linear transformation matrix to project them into another monolingual space with very little supervision i.e requiring a \textit{weak cross-lingual signal} \cite{artetxe2018aaai, mikolov2013exploiting} or 2) \textit{joint methods} which jointly optimise monolingual as well as cross-lingual learning objectives using parallel corpora thus requiring a \textit{strong cross-lingual signal}. \cite{gouws2016bilbowa, luong2015bilingual, lample2018phrasebased}

\noindent As mapping methods use transformation matrices to align embedding spaces they make the crucial assumption that, regardless of domain or linguistic differences, these spaces are \textit{approximately isomorphic} i.e. they share a similar structure \cite{vulic2020good}. \footnote{Two vector spaces are said to be isomorphic if there is an invertible linear transformation from one to the other.} It has been shown \cite{sogaard2018limitations, vulic2020good} that this assumption does not hold in general and therefore the benefit of mapping methods requiring little to no cross-lingual signal can no longer be taken advantage of directly in low-resource scenarios. At the same time, while joint methods do not make the isomorphism assumption \cite{ormazabal-etal-2019-analyzing} they are inapplicable in low-resource settings due to their high data requirements. While most recent work in low-resource CLWEs have focused on reducing the supervision signal as much as possible \cite{artetxe2018unsupervised}, further study points to this not being the best approach \cite{vulic2019really}. We claim that in addition to utilising monolingual resources, related language parallel data can be crucial in artificially generating isomorphic embedding spaces between the source and target.

\noindent In this paper, we address the limitations outlined above by proposing an alternative method to learn CLWEs for low-resource and distant language pairs. Contrary to previous approaches, we combine the benefits from both mapping and joint-training methods to develop high-quality, isomorphic embeddings. In our proposed framework, we maintain the low level of supervision as obtained by mapping methods while still guarding the isomorphic embeddings achieved by joint-training by independently aligning source and target embeddings to a related higher-resource language. We apply our method in several low-resource settings and conduct evaluations on bilingual lexicon induction and eigenvalue similarity. Our experiments show that, despite no additional source-target parallel data, our approach outperforms conventional mapping and joint-training methods on both evaluation metrics. 

\noindent The main contributions of this work can be outlined as the following:

\begin{itemize}
    \item We introduce a framework combining mapping and joint methods to learn isomorphic cross-lingual embeddings for low-resource language pairs.
    \item We successfully employ CLWEs in challenging, low-resource scenarios without the use of explicit source-target parallel data.
    \item We achieve significant gains over state-of-the-art methods in both bilingual word induction as well as eigenvalue similarity.
\end{itemize}

\section{Related Work}

\paragraph{Cross-Lingual Word Embeddings} \noindent CLWEs aim to represent words from several languages into a shared embedding space which allows for several applications in low-resource areas such as transfer learning \cite{peng2021summarising}, NMT \cite{artetxe2018unsupervised}, and Bilingual Lexicon Induction (BLI) \cite{patra-etal-2019-bilingual}. Largely, there are two classes of approaches to learn CLWEs: \textbf{mapping} and \textbf{joint} methods. While the former aims to map monolingually learnt embeddings together, the latter simultaneously learns both embedding spaces using some cross-lingual supervision (i.e. a cross-lingual signal). Common approaches to achieve this cross-lingual signal come from parallel corpora aligned at the word \cite{luong2015bilingual} or sentence level \cite{pmlr-v37-gouws15}. In addition to this, later methods proposed the use of comparable corpora \cite{vulic2016bilingual} or large bilingual dictionaries \cite{duong-etal-2016-learning} as a form of supervision. For a more detailed survey of methods and limitations of CLWEs, the reader is referred to \cite{ruder2019survey}.

\paragraph{Offline Mapping}\noindent As mapping methods map monolingual embedding spaces together, instead of relying on a cross-lingual signal (such as in joint methods) they work by finding a transformation matrix that can be applied to the individual embedding spaces. In the case of supervised learning, a large bilingual dictionary would have been used as supervision however \cite{artetxe2018acl} get rid of this requirement via a self-learning strategy. Their approach is based on a robust iterative method combined with initialisation heuristics to get state-of-the-art performance using offline mapping. Most of these methods align spaces using a linear transformation- usually imposing orthogonality constraints- in turn assuming that the underlying structure of these embeddings are largely similar. Several works \cite{sogaard2018limitations, vulic2020good} have shown that this assumption does not hold when working with non-ideal scenarios such as low-resource or typologically different language pairs. In order to mitigate this assumption, \cite{mohiuddin-etal-2020-lnmap} learn a non-linear map in a latent space, \cite{nakashole-2018-norma} uses maps that are only locally linear, and \cite{glavas-vulic-2020-non} propose to learn a separate map for each word. However these are supervised methods, meaning they suffer from limitations of hubbness and isomorphism as outlined in \cite{ormazabal-etal-2019-analyzing}. To address these limitations, \cite{ormazabal2021offline} proposes a method in which they fix the target language embeddings, and learn a new set of embeddings for the source language that are aligned with them using self-learning. Their method outperforms current mapping, joint, as well hybrid methods on the MUSE dataset \cite{conneau2018word}. 

\paragraph{Joint-Training}\noindent The fundamental limitations of offline methods are not faced by joint-training methods if there is a strong cross-lingual signal available \cite{ormazabal-etal-2019-analyzing}. In practice, however, we don't always have access to such forms of supervision therefore recent works have attempted to reduce the supervision level so as to preserve the isomorphism achieved by joint methods while still being as widely applicable as mapping methods. \cite{lample2018phrasebased} use concatenated monolingual corpora in different languages and learn word embeddings over this constructed corpus, using identical words as anchor points. \cite{devlin2019bert} use a bidirectional transformer to learn a multilingual embedding space which showed significant progress in the zero-shot cross-lingual transfer task. Further extending upon these works, \cite{wang2020crosslingual} effectively combined joint and mapping based methods in their framework {``}joint-align'' however their method was not tested on distant language low-resource pairs. In their work, they use fully unsupervised joint initialisation as the first step, vocabulary reallocation where they {``}unshare'' some vocabulary to better align them, and lastly they perform a refinement step using off-the-shelf alignment methods. Furthermore, for low-resource setups \cite{kementchedjhieva-etal-2018-generalizing} propose Multi-support Generalized Procrustes Analysis (MGPA) which learns a three-way alignment between English, a low-resource language, and a supporting related language. In addition to \cite{wang2020crosslingual}, \cite{woller-etal-2021-neglect} show the benefit of using related languages in the context of CLWEs. In their work, they use a two step approach where they first use joint-align to learn a CLWE between the low-resource and related language and as a next step they map it to the target language to build the final multilingual embedding space. Although they focus on the use case of Occitan via French, Spanish, and Catalan, we show that this type of approach is well motivated across several language pairs.

\section{Methodology}

\begin{figure}[htp]
    \centering
    \includegraphics[width = 0.4\textwidth]{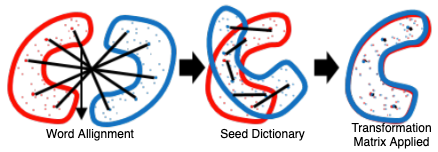}
    \caption{Toy visualisation of mapped cross lingual embedding spaces with red representing one language and blue the other}
    \label{fig:maps}
\end{figure}

\noindent Given two embedding spaces, $X$ and $Y$, for languages $x$ and $y$ respectively, our goal is to align them together without any direct parallel data between them and without assuming orthogonality/structural similarity. In order to do this, let us consider a third embedding space, $Z$, of a language $z$ related to the source $x$. Furthermore, let there also be sufficient parallel data between $y$ and $z$ to jointly learn their aligned embedding spaces \cite{ormazabal-etal-2019-analyzing}. Our approach first aligns the spaces $X$ and $Z$ using an unsupervised offline mapping method \cite{artetxe2018acl}. \cite{vulic2020good} find that for typologically similar languages that have in-domain monolingual corpora, isomorphism in their learnt vector spaces is preserved. To that end, due to the linguistic similarities between $x$ and $z$ we may perform offline mapping. Figure \ref{fig:maps} shows a visualisation of how these two embedding spaces are aligned using an induced seed dictionary as per \cite{artetxe2018acl}. For further details about the offline alignment, the reader is referred to read the original paper. \\

\begin{figure}[htp]
    \centering
    \includegraphics[width = 0.5\textwidth]{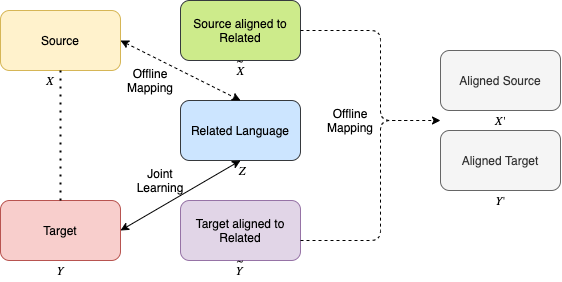}
    \caption{Visualisation of our proposed alignment method in context; dotted lines represent lack of parallel data between language pairs}
    \label{fig:flow}
\end{figure}

\noindent Once the space $X$ is aligned to the monolingual space $Z$, we wish to also align $Y$ to $Z$ as well. Due to the typological differences between the two languages, we can no longer assume isometry of their embedding spaces therefore can no longer use offline mapping methods. However, due to higher-resource nature of $z$, we have access to parallel corpora between $y$ and $z$. This allows us to apply joint-training approaches \cite{luong2015bilingual} to simultaneously learn their embeddings. As found in \cite{ormazabal-etal-2019-analyzing}, under ideal conditions of having parallel data, joint-training approaches produce isomorphic embeddings that perform better than their offline counterparts in bilingual lexicon induction. As shown in Figure \ref{fig:flow}, we can now produce two embedding spaces, Source aligned to Related and Target aligned to Related while preserving isomorphism. As a final step in our alignment framework, we use the $z-$aligned embedding spaces, $\tilde{X}$ and  $\tilde{Y}$, to induce the final cross-lingual word embedding spaces. Now that both $X$ and $Y$ are projected onto $Z$, they share structural similarity which permits the use of offline mapping on $\tilde{X}$ and  $\tilde{Y}$. Figure \ref{fig:flow} shows the complete alignment framework to produced the resultant isomorphic embedding spaces. 

\noindent Our proposed framework can be summarised in the following steps:

\begin{enumerate}
    \item For a source-target pair, choose a related higher-resource language to the low-resource target such that there is \textbf{sufficient} source-related parallel data to perform joint mapping. \cite{ormazabal-etal-2019-analyzing}
    \item Use offline mapping \cite{artetxe2018acl} to align related and source language into a shared embedding space. Due to their relatedness, these resultant embeddings remain isomorphic as the assumption in mapping methods hold true.
    \item Use joint training \cite{luong2015bilingual} to map related and target language into a shared embedding space using the higher-resource parallel data between them. As this is the highest level of supervision possible, we ensure that the embedding spaces remain isomorphic.
    \item Lastly, map the aligned-source and aligned-target embeddings using unsupervised mapping methods as they are now isometric in nature following the alignment to the related language for both the source and target. 
\end{enumerate}
\noindent This framework uses the low cross-lingual signal utilised by mapping techniques while still maintaining the isomorphism of the resultant embedding spaces as in joint approaches. This is achieved by exploiting the existing isomorphism between embeddings as much as possible by pre-aligning the spaces via a pivot-language. However, unlike pivot-based MT we do not compound errors across embedding spaces due to the final refinement step done by mapping the aligned embeddings into their shared cross-lingual space \cite{10.1145/3406095}. The first step of cross-lingual mapping, allows us to internalise the structure of the low-resource embedding space by pre-aligning with the related language. In the second step, we re-learn a joint embedding space for the related language and the target. In the last stage, the offline mapping makes use of the internalised structure by associating the modified source embedding space with the modified target embedding space which have both independently been aligned to the same related language. \\

\noindent With this pipeline, we are able to target a large group of low-resource languages which belong to higher-resource language families for instance, English-Nepali via Hindi. Linguistically, Nepali and Hindi are quite similar as they share the same script and also have 80\% of subword tokens in common when using a shared BPE vocabulary of 100k subword units \cite{lample2019crosslingual}. In this work, we perform experiments on several low-resource language pairs to show the effectiveness of our approach in various language families- specifically we look at Uralic, Indo-European, and Romance languages. \\

\noindent Our goal is not to fully replace  current methods of learning cross-lingual word representations but to aid them in the area of low-resource languages. As shown by \cite{ormazabal-etal-2019-analyzing}, depending on the type of resources available as well as the languages considered, different methods can be preferred. While current approaches perform well for several languages and resource levels \cite{ormazabal2021offline}, their performance still leaves room for improvement in the low-resource, typologically diverse area. Despite the simplicity of our method, our experiments show that we perform competitively on quality as well as degree of isomorphism across all low-resource pairs considered. Due to the reliance on a sufficiently resourced related language, our method is not applicable to every low-resource pair however referring to the task of related-language NMT we see that there is indeed a large group of languages that could benefit from this approach \cite{10.1145/3406095}.

\section{Experimental Design}

In this section we discuss the datasets used, training settings for different configurations used in our experiments, and lastly the evaluation metrics used to assess the embedding spaces produced by our framework.

\subsection{Datasets}

\noindent In our work, we train CLWEs between English and five other low-resource languages: Nepali (ne), Finnish (fi), Romanian (ro), Gujarati (gu), and Hungarian (hu). We use pre-trained fasttext word embeddings \cite{grave2018learning} which uses Wikipedia dumps and Common Crawl for all languages. In addition to this, we use available parallel data between the following related language pairs respectively: English-Hindi (hi) for Nepali, English-Estonian (et) for Finnish,  English-Italian (it) for Romanian, English-Hindi (hi) for Gujarati, and English-Finnish (fi) for Hungarian. We obtain the data from IIT Bombay\footnote{\url{http://www.cfilt.iitb.ac.in/iitb_parallel/}} for En-Hi and from the WMT workshops.\footnote{\url{http://www.statmt.org}} We preprocess all the data using Moses scripts and tokenise using BPE. For the Indic languages, we use IndicNLP \footnote{\url{https://github.com/anoopkunchukuttan/ indic_nlp_library}} for word segmentation. Table 1 details the statistics of the approximate corpus sizes used for learning monolingual embeddings. For evaluation, we use the gold-standard bilingual dictionary from the MUSE dataset \cite{conneau2018word} for Finnish and for the remaining language pairs, we use bilingual dictionaries published by \cite{pavlick-etal-2014-language}. To show the relatedness of the languages chosen, we report genetic proximities \footnote{\url{https://www.elinguistics.net/Compare_Languages.aspx}} of the pairs with their related languages in Table \ref{tab:langpairs}.

\begin{table}[htp]
\centering
\begin{tabular}{@{}lrr@{}}
\hline
 &\textbf{Sentences} & \textbf{Tokens}\\
 \textbf{Languages} & & \\
\hline
Ne & 92.3K & 2.8M\\
Fi & 6M & 91M \\
Ro & 88.6K & 2.28M \\
Gu & 382K & 6M \\
Hu & 1M & 15M \\
En & 67.8M & 2.0B \\
\hline
\end{tabular}
\caption{Monolingual Training Corpora sizes}
\label{tab:data}
\end{table}

\begin{table}[htp]
\centering
\begin{tabular}{@{}lr@{}}
\hline
 & \textbf{Segments}\\
 \textbf{Language Pairs} & \\
\hline
Hi-En & 1.5M\\
Et-En & 1.7M \\
It-En & 151M \\
Fi-En &  6.2M\\
\hline
\end{tabular}
\caption{Parallel Training Corpora sizes}
\label{tab:pardata}
\end{table}

\begin{table}[htp]
\centering
\begin{tabular}{@{}lllr@{}}
 \textbf{x} & \textbf{z} & \textbf{y} & Gen. Proximity ($\downarrow$) \\
\hline
Nepali & Hindi & En & 19.4 \\
Finnish & Estonian & En & 16.7\\
Gujarati & Hindi & En & 31.8\\
Romanian & Italian & En & 29.4\\
Hungarian & Finnish & En & 62.2\\
\hline
\end{tabular}
\caption{Language Set-Ups with Genetic Proximity of Source and Related language where lower is better}
\label{tab:langpairs}
\end{table}

\subsection{Training Settings}

\noindent \textbf{Mapping:} Using fasttext \cite{grave2018learning} with the default parameters \footnote{These are 300-dimensional vectors with 10 negative samples, a sub-sampling threshold of 1e-5 and 5 training iterations}, we first gather monolingual word embeddings for each of the respective languages. After this, we map the embeddings to a cross-lingual space using VecMap \cite{artetxe2018acl} in the \textbf{unsupervised mode} as we do not have any bilingual dictionaries. In this mode an initial solution is found using heuristics and iteratively refined. \\

\noindent \textbf{Joint Training:} To train the embeddings jointly, we use the BiVec tool proposed by \cite{luong2015bilingual} which is an extension of skip-gram algorithm aiming to predict the context around both the source and target word aligned to a given parallel corpus at the word level. We use the same hyperparameters as in the mapping methods. \\

\noindent In addition to the mapping and joint-training methods trained as described earlier, we also train Joint Align \cite{wang2020crosslingual}. In order to this, we use the official implementation \footnote{\url{https://github.com/thespectrewithin/joint_align}} on preprocessed tokenised data. We use the non-contextual model in specific as we are working on non-contextual word embeddings. Furthermore, we replace the RCLS retrieval step with unsupervised VecMap to have more consistency across the baselines. All other baselines are trained with best settings as specified in the respective papers. 

\subsection{Evaluation Metrics}

\noindent We evaluate our embeddings  on two aspects: the quality and degree of isomorphism achieved between the source and target. As in \cite{ormazabal-etal-2019-analyzing}, we measure this by bilingual lexicon induction (BLI) and eigenvalue similarity respectively. Firstly, we induce the word-level translations by linking neighbouring source-target word translations in the resultant embeddings spaces using CSLS retrieval and finally evaluate the induced dictionary against the English-Target bilingual dictionary released by \cite{pavlick-etal-2014-language} to compute precision scores for the BLI task using the MUSE evaluation scripts \cite{conneau2018word}. \footnote{\url{https://cs.brown.edu/people/epavlick/data.html}} Next, we measure eigenvalue similarity for the embeddings following the procedure in \cite{sogaard2018limitations} on centralised and normalised embeddings. We perform the same evaluations across different cross-lingual alignment methods on all the considered language pairs.

\section{Results and Discussion}

\noindent In this section, we discuss our main experimental results on BLI and eigenvalue similarity across the chosen language pairs. Furthermore, we also conduct ablation tests on our learnt embeddings to further verify the sources of improvements. 
\subsection{BLI}
\begin{table*}[htp]
    \centering
    \begin{tabular}{lllllll}
         & ne $\rightarrow$ en & fi $\rightarrow$ en & ro $\rightarrow$ en & gu $\rightarrow$ en & hu $\rightarrow$ en  & \textbf{avg}\\
         \bottomrule 
         VecMap \cite{artetxe2018acl}& 52.3 & 61.9 & 61.6 & 45.4 & 53.2  & 54.8 \\
         Joint \cite{luong2015bilingual}& 21.3 & 30.5 & 31.4 & 33.4 & 25.5 & 28.4 \\
         Joint Align \cite{wang2020crosslingual}& 24.5 & 31.3 & 28.2 & 35.4 & 26.5  & 25.2  \\
         MGPA \cite{kementchedjhieva-etal-2018-generalizing} & 41.6 & 55.7 & 57.3 & 39.6 &45.7 & 47.9 \\
         Joint Align + MUSE \cite{woller-etal-2021-neglect} & \textbf{59.4} & 62.5 & 62.6 & \textbf{49.1} & 55.8 & 57.8 \\
         Ours & 58.4 & \textbf{65.2} & \textbf{64.5} & 48.4 & \textbf{56.3} &  \textbf{58.6}\\  
    \end{tabular}
    \caption{Precision at 1 scores of proposed method and previous works on BLI (higher is better)}
    \label{tab:bli}
\end{table*}

Results in Table~\ref{tab:bli} report the BLI scores for the different baselines and our proposed method. We use the Low-Resource to English language direction however MGPA can only be trained with the related and low-resource language at the target. As per \cite{woller-etal-2021-neglect}, we evaluate the resultant Low-Resource-English embeddings afterwards. Across all language pairs, we see substantial gains from our method as compared to mapping, joint, and other hybrid baselines. \cite{woller-etal-2021-neglect} outperform our approach on two language pairs (Ne-En, Gu-En) which we suspect is due to Joint-Align's performance in comparison to regular Joint training. However, these improvements are not consistent. As Joint-Align uses a vocabulary re-sharing step, we can hypothesise that for language pairs with significant vocabulary overlap this step might be useful in learning better alignments. In particular, 
Joint Align on average
performs poorly on most 
language pairs, suggesting that it is inapplicable in a truly low-resource scenario. 
VecMap 
performs well overall,
however, our approach outperforms VecMap by a significant margin 
. Despite using VecMap and a purely joint-training based approach without any additional source-target supervision, the gains in the scores are substantial. Interestingly, our method performs well even in the case of fi $\rightarrow$ en where we use Estonian as the related language;  
Estonian is in fact lower-resource than Finnish, however our performance suggests that "pivoting" via Estonian was still helpful in learning Finnish-English word embeddings. Therefore, even if the embeddings learnt in the intermediate stages are not ideal, 
the structural alignments earned are ultimately helpful in obtaining better source-target embeddings. 

\subsection{Eigenvalue Similarity}
\begin{table*}[htp]
    \centering
    \begin{tabular}{lllllll}
         & ne $\rightarrow$ en & fi $\rightarrow$ en & ro $\rightarrow$ en & gu $\rightarrow$ en & hu $\rightarrow$ en  & \textbf{avg}\\
         \bottomrule 
         VecMap \cite{artetxe2018acl}& 205.8 & 118.2 & 176.4 & 189.3 & 94.5 & 156.8\\
         Joint \cite{luong2015bilingual}& 48.6 & 30.3 & 41.2 & 42.5& 35.6 &  39.7\\
         Joint Align \cite{wang2020crosslingual} & 56.2& 45.5 & 50.1 & 48.2 & 38.6 & 47.7 \\
         MGPA \cite{kementchedjhieva-etal-2018-generalizing} & 58.4 & 48.1 & 48.3 & 52.5 & 35.1 & 48.8 \\
          Joint Align + MUSE \cite{woller-etal-2021-neglect} & 38.3 & 28.1 & 35.6 & 37.1 & 28.1 & 33.4 \\
         Ours & \textbf{37.5} & \textbf{23.4} & \textbf{32.7} & \textbf{33.2} & \textbf{26.6} & \textbf{30.7}\\  
    \end{tabular}
    \caption{Eigenvalue Similarity Scores (lower is better)}
    \label{tab:eigsim}
\end{table*}

In eigenvalue similarity, mapping methods perform much worse than joint training
(Table \ref{tab:eigsim}).  
This finding is in line with the literature \cite{ormazabal-etal-2019-analyzing}, and is explained by the high linguistic divergence between English and source languages, resulting in embeddings that are far less isomorphic. Our hybrid approach performs even better than joint methods and achieved the best eigenvalue similarity score across all language pairs, showing that we do indeed obtain isometric embeddings while still not requiring the higher level of supervision in joint learning approaches. Although our proposed framework does not make any significant changes to the mapping and joint components, the combination of the two cross-lingual approaches leads to better embeddings both in terms of quality, shown by the performance in BLI, as well as structure, shown by the eigenvalue similarity scores. In addition to this, MGPA as well as \cite{woller-etal-2021-neglect}'s method attains good eigenvalue similarity scores suggesting that the incorporation of a related language is indeed helpful 

\subsection{Ablation Tests} 

\noindent To study where the improvements of the cross-lingual encoding method come from, we conduct several ablation tests  
(results in Table \ref{tab:schemes}), 
assessing the contribution of different embedding schemes to the final quality of the embeddings: firstly, we look at the initial unaligned monolingual embeddings, next we look at the embeddings that are independently aligned to the related language, and lastly we look at the embeddings after the final offline map has been constructed.  These embedding schemes allows us to verify the importance of the intermediate structural alignments via the related language. As expected the unaligned embeddings have a near 0 BLI score, suggesting that the initial embeddings do not have any linking however as the score is still non-zero we can attribute this to identical words across some language pairs. However, the intermediate embeddings obtained (Related-Aligned in Table \ref{tab:schemes}) have a significant jump in performance even though there is no explicit alignment between the source and target at this stage. This intermediate performance is surprisingly close to the final performance obtained by Joint Align as well, which suggests that the related-language strategy allows for a better understanding of word associations even before performing the final step of offline mapping. 
\begin{table*}[htp]
    \centering
    \begin{tabular}{lrrrrrrrrrrrrrrrrrr}
            &
         \multicolumn{3}{c}{ne $\rightarrow$ en} &  \multicolumn{3}{c}{fi $\rightarrow$ en} &  \multicolumn{3}{c}{ro $\rightarrow$ en} &  \multicolumn{3}{c}{gu $\rightarrow$ en} &  \multicolumn{3}{c}{hu $\rightarrow$ en} \\
& hi & gu & et & et & hu & hi & it & fr & de & hi & ta & fr & fi & et & fr \\
         \textbf{Rel} & 24.7 & 21.3 & 15.8 & 33.4 & 27.1 & 19.6 & 33.8 & 30.7 & 21.3 & 24.6 & 16.8 & 14.6 & 23.9 & 22.8 & 13.6\\
         \textbf{Full} & 58.4 & 42.3 & 30.5 & 65.2 & 54.3 & 38.1 & 64.5 & 61.3 & 42.6 & 48.4 & 32.3 & 29.6 & 56.3 & 55.3 & 36.7\\
         \bottomrule 
    \end{tabular}
    \caption{BLI Scores P@1 for different related languages }
    \label{tab:relab}
\end{table*}

\noindent Next, we studied the relevance of the relatedness as well as amount of parallel language of the related language. For this ablation test, we took three languages of different degrees of relatedness to each source language and then we measured the improvements between the intermediate embedding space aligned to the related language and the final embedding space between the source and the target. Doing so allowed us to further isolate intermediate improvements as obtained by the related languages and their final contribution in the quality of the learnt embeddings. We report results in Table \ref{tab:relab}. Consistently, across all the language pairs considered the BLI scores take a sharp drop as we reduce the relatedness (as measured by lexical similarity) of the intermediate language. The scores here point to relatedness of the chosen language as being one of the key factors in improving downstream performance. These results are in line with findings from \cite{woller-etal-2021-neglect} where the relatedness of Catalan to Occitan was the driving force in their performance even though the resource levels of French and Spanish were significantly higher than Catalan.

\begin{table}[htp]
\centering
\begin{tabular}{lr}
\hline
 & BLI Score\\
 \textbf{Embeddings} & \\
\hline
\multicolumn{2}{c}{Our Method} \\
\hline
Unaligned & 0.4 \\
Related-Aligned & 24.6\\
Full Alignment & 58.6 \\
\hline
\multicolumn{2}{c}{Offline Mapping} \\
\hline
Unaligned & 0.4 \\
Mapped & 54.8\\
\hline
\multicolumn{2}{c}{Joint Align} \\
\hline
Unaligned & 0.4 \\
Aligned & 25.2\\
\hline
\end{tabular}
\caption{Ablation Tests on Different Embeddings, reporting average Precision @ 1 score}
\label{tab:schemes}
\end{table}

\section{Conclusion and Future Work}

\noindent In this work, we developed a framework to learn cross-lingual word embeddings in low-resource scenarios. We addressed limitations of both offline as well as joint training methods to develop high quality, isomorphic embeddings for several low-resource language pairs. In particular, we maintain the low cross-lingual signal as required by offline methods while still obtaining structurally sound/isomorphic embeddings as in joint-training based approaches. Our method works by exploiting a higher-resource related-language to jointly learn a cross-lingual space between the related-language and target while also learning a cross-lingual space between the source and the related language using offline mapping. Due to the pre-alignment with a related-language, the resultant cross-lingual spaces are now structurally similar and can be mapped to each other without breaking any orthogonality assumption. Whilst our approach does not change the individual components at all, we obtain far superior results in both BLI as well as eigenvalue similarity across all languages. On a high-level, the gains in our method can be attributed to incorporating more linguistic information in the low-resource language via the related language. This would in turn allow for better modelling of the structure of the embedding spaces without explicitly requiring additional source-target parallel data. As our ablation tests show, indeed the intermediate embeddings themselves have some performance gains even though the source and target embeddings are not aligned to each other yet. 

\noindent Future work in this direction would include verifying how high-resource the related language needs to be to still see performance gains. In addition to this, we would like to explore how the relatedness of the pivot language affects the performance of the learnt embeddings. Specifically, we would like to discover to what extent isomorphism is preserved in related language pairs- permitting the use of offline methods in more distant languages. Studying this would allows us to suggest further generalisations of our approach to cover a wider range of language families. 

\bibliography{custom}
\bibliographystyle{acl_natbib}

\end{document}